\documentclass[12pt]{iopart}

\usepackage{color}
\usepackage{graphicx}
\expandafter\let\csname equation*\endcsname\relax
\expandafter\let\csname endequation*\endcsname\relax
\usepackage{amsmath}
\usepackage{amssymb}
\usepackage{mathptmx} 
\usepackage{mathtools}
\usepackage[ruled]{algorithm2e}

\DeclareMathOperator*{\argmin}{arg\,min}

\begin{document}

\title{Optimisation via encodings: a renormalisation group perspective}

\author{Konstantin Klemm}
\address{IFISC (CSIC-UIB), Campus Universitat de les Illes Balears, 
    E-07122 Palma de Mallorca, Spain}
\ead{klemm@ifisc.uib-csic.es}

\author{Anita Mehta}
\address{
  Bioinformatics Group, Department of Computer Science and
  Interdisciplinary Center for Bioinformatics,
  University Leipzig, D-04107 Leipzig, Germany}
\address{Faculty of Linguistics, Philology and Phonetics,
  Clarendon Institute, Walton Street,
  Oxford OX1 2HG, UK}
\address{Max Planck Institute for Mathematics in the Sciences,
  Inselstrasse 22, D-04103 Leipzig, Germany}
\ead{anita@bioinf.uni-leipzig.de}

\author{Peter F. Stadler}
\address{
  Bioinformatics Group, Department of Computer Science and
  Interdisciplinary Center for Bioinformatics,
  University Leipzig, D-04107 Leipzig, Germany}
\address{
  Max Planck Institute for Mathematics in the Sciences,
  Inselstrasse 22, D-04103 Leipzig, Germany}
\address{Santa Fe Institute, Santa Fe, NM 87501, USA}
  \ead{studla@bioinf.uni-leipzig.de}

\begin{abstract}  
  Difficult, in particular NP-complete, optimization problems are
  traditionally solved approximately using search heuristics.  These are
  usually slowed down by the rugged landscapes encountered, because
    local minima arrest the search process.  Cover-encoding maps were
  devised to circumvent this problem by transforming the original landscape
  to one that is free of local minima and enriched in near-optimal
  solutions.  By definition, these involve the mapping of the original
  (larger) search space into smaller subspaces, by processes that typically
  amount to a form of coarse-graining. In this paper, we explore the
  details of this coarse-graining using formal arguments, as well as
  concrete examples of cover-encoding maps, that are investigated
  analytically as well as computationally.  Our results strongly suggest
  that the coarse-graining involved in cover-encoding maps bears a strong
  resemblance to that encountered in renormalisation group schemes. Given
  the apparently disparate nature of these two formalisms, these strong
  similarities are rather startling, and suggest deep mathematical
  underpinnings that await further exploration. \newline
Published as J Phys A: Math Theor 56, 485001 (2023)
\end{abstract}

\vspace{2pc}


\maketitle

\section{Introduction} 
\label{sect:intro}

There has been a body of work on the optimisation of complex problems, in
particular to do with NP-complete problems. These are typically approached
by algorithms utilizing local search which, in analogy with the field of
evolutionary biology, are often described as dynamics on a fitness
landscape (see, e.g.\ the book by Engelbrecht and Richter
\cite{Engelbrecht:14}).  The performance of local search algorithms depends
strongly on the structure of the search spaces involved. These in turn are
determined by two largely independent ingredients: (1) the concrete
representations of the configurations that are to be optimized, referred to
as encodings, and (2) locality in search space, usually determined by a
move set.

For many well-studied combinatorial optimization problems and related
models from statistical physics (such as spin glasses), there is a natural
encoding. For instance, the tours of a Travelling Salesman Problem (TSP)
are naturally encoded as permutations of the cities concerned, while spin
configurations are encoded as strings over the alphabet $\{+,-\}$ with each
letter referring to a fixed spin variable. This natural encoding is usually
free of redundancy; any residual redundancies that occur usually arise from
simple symmetries of the problem that can be factored out easily. For
instance, tours in the Traveling Salesman Problem \cite{Applegate:06} can
start at any city -- therefore they are invariant under rotations. Many
spin glass models are invariant under the simultaneous flipping of all
spins.

For a given encoding, the performance of search crucially depends on the
move set. Here, we will consider only reversible, mutation-like moves. The
search space is therefore modeled as an undirected graph.  The cost
function assigned to a specific search space defines a \emph{fitness
landscape}. Evolutionary algorithms can thus be viewed as dynamical systems
operating on landscapes, whose structure has, as a consequence, been
studied extensively in the field
\cite{Reidys:02a,Ostman:10,Engelbrecht:14}.

Such approaches are however limited by the slowing down that occurs when
the local minima that are a feature of the typically rugged landscapes
encountered, arrest the progress of the search process. Another way of
tackling optimization problems is by the use of heuristic approximations to
estimate a global cost minimum. In \cite{Klemm:18a}, we presented a
combination of these two approaches by using \emph{cover-encoding maps},
which map processes from a larger search space to subsets of the original
search space. The key idea was to construct cover-encoding maps, with the
help of suitable heuristics, that singled out near-optimal solutions, and
resulted in landscapes on the larger search space that no longer exhibited
trapping local minima.
  
One of the features that emerged when this was done was that the maps
typically involved a process of coarse-graining. Further considerations
suggested that in fact there were striking similarities between these
optimisation schemes and the schemes that are normally associated with the
renormalisation group in physics.

The main aim of this paper is to use different examples of NP-complete
problems to illustrate the analogies between the construction of our
cover-encoding maps and the coarse-graining involved in renormalisation
group schemes.  The plan of our paper is as follows. In Section 2, we first
revisit the basics of our theory of cover-encoding maps \cite{Klemm:18a},
with particular reference to the Number-Partitioning Problem (NPP) which
inspired it. We next use the quadratic spin glass to provide a concrete
illustration of the coarse-graining inherent in the construction of
cover-encoding maps, via detailed explorations which are analytical as well
as computational.  In Section 3, we move to the more formal aspects of the
analogies between cover-encoding maps and renormalisation group schemes,
involving a step-by-step comparison of their structure as well as their
flow properties. In the Discussion section, we provide a succinct account
of the similarities and differences between these two kinds of
coarse-grainings, and discuss the open questions unravelled by our work.

\section{Theory of cover-encoding maps: application to a quadratic spin glass}

The initial motivation for the construction of cover-encoding maps was the
decades-old observation that certain redundant encodings of the
Number-Partitioning Problem (NPP) allow simple, generic optimization
heuristics to find dramatically improved solutions \cite{Ruml:1996}.  In
previous work \cite{Klemm:12b} we found that this approach was not limited
to the NPP, but that suitably chosen redundant encodings also improved the
performance of heuristics on several other combinatorial optimization
problems. In more recent work \cite{Klemm:18a}, we demonstrated (a) why the
particular method used by \cite{Ruml:1996} works so well and (b) how it can
be generalized to arbitrary combinatorial optimization problems.

Our focus in \cite{Klemm:18a} was on black-box-type optimization scenarios in
which the information on the cost function $f(x)$ is exclusively obtained
by evaluating it for specific configurations $x\in X$ in the search space
$X$. The sequence of these function evaluations is determined by the
optimization heuristic. Practical algorithms of this type propose
candidates $x\in X$ for evaluation based on past evaluation results. These
candidates are chosen locally in the vicinity of past successful candidates
with the help of rules that depend on the representation of $X$.  This
explicitly or implicitly defines a topological structure on $X$. For the
purpose of the present contribution, we assume that the topology of the
search space $X$ is expressed by a notion of adjacency that is respected by
the search process.

Intuitively, the most important obstruction for local optimization
heuristics is the presence of a large number of local optima that trap the
search process. The aim of a redundant encoding, therefore, is to provide
an alternative representation $Y$ of the optimization problem that reduces
the number of local optima and makes it easier to find the globally optimal
solution. Such a formulation $Y$ should, ideally, be such that:
\begin{itemize} 
\item[(i)] neighborhoods in $Y$ are small enough to be searched in
  practice.
\item[(ii)] for every starting point there is a path to the global optimum 
  such that the cost function is decreasing, or at least non-increasing.
\end{itemize}
Condition (i) ensures that we still deal with local search heuristics,
while condition (ii) intuitively makes the landscape easy to search. Note
that condition (ii) does not make the optimization problem trivial, since
the heuristics still have to find an efficient path among possibly many
very long ones. Its real significance is that it rules out traps and
guarantees that simple downhill search will, eventually, be successful.

The implementation of the above conditions led to the construction of
cover-encoding maps, whose theory will be presented in the next subsection.

\subsection{General theory of Cover-encoding Maps} 

The key idea of cover-encodings is to replace the original optimization
problem by a collection of ``coarse-grained'' optimization problems that
collectively still cover all ``relevant'' solutions of the original
problem. As we shall see below, the notion of coarse-graining a problem is
intuitively appealing but difficult to grasp precisely in mathematical
terms. Previously, we considered \emph{restrictions} as an alternative
\cite{Klemm:18a}. As we shall argue, these restrictions can often (but not
necessarily always) be chosen such that they conform to the idea of
coarse-graining.

We choose  the Number Partitioning Problem (NPP)
\cite{Garey:1979a} as a simple and instructive example of cover-encoding maps
and coarse-graining.
An NPP instance of size $N$ is given by $N$ positive numbers
$(a_1,a_2,\dots,a_N)$. The optimization problem is to divide these numbers into
two subsets such that the sums over subsets are as close as possible to each
other. This amounts to finding a vector of $\pm$ signs (or binary spins) 
$s=(s_1,s_2,\dots,s_N) \in \{-1,+1\}^N$ to minimize
$ f(x) = | \sum_{i=1}^n s_i a_i | $.

Cover encodings will be derived from restricted sets of solutions.  For the
NPP, a \emph{prepartitioning} \cite{Ruml:1996} assigns a cluster $y_i$ to
each index $i\in\{1,\dots,N\}$ such that $a_i$ and $a_j$ go into the same
subset if $i$ and $j$ are in the same cluster. In terms of the sign vectors
$s$, the prepartioning restricts the solution space to the solutions with
$s_i = s_j$ when $y_i=y_j$. Typically all cluster indices are chosen from
the range $y_i \in \{1,2,\dots,N\}$ allowing to express the unconstrained
original problem by $y = (y_1,\dots,y_N) = (1,\dots,N)$ or a permutation
thereof. As the other extreme, restriction to a single specific solution
$(s_1,\dots,s_N)$ is encoded by the prepartitioning with $y_i = 1$ for
$s_i=-1$, and $y_i=2$ for $s_i=+1$.  The role of the prepartitioning
algorithm is to enable global as well as local sampling: elements, be they
distant or close on the original string, can be grouped together if they
have the same prepartioning index -- a feature which is likely responsible
\cite{Klemm:18a} for the superior performance of this scheme compared to
the others in \cite{Ruml:1996}.

The optimization problem in the solution space restricted by a
prepartitioning $y$ is again an instance of NPP, generally of smaller size
$N'$ and with transformed numbers. These are obtained by summing over the
numbers in each non-empty cluster $c$, $a'_c = \sum_{\{i:y_i=c\}}
a_i$. Thus $y$ also defines a coarse-grained instance of NPP, whose input
consists of one number for each class $c$ of the partition $y$.  The
coarse-grained problem solves the original problem provided $y$ is chosen
so that it is a refinement of the optimal bipartition $y^*$ that solves the
NPP. Note that $y^*$ is also a restriction of the original problem.

In the general case of a cost function $f:X\to\mathbb{R}$ over an arbitrary
(finite) domain $X$, the analog of prepartitioning is to define a
collection $\{\varphi(y)|y\in Y\}$ of subsets of $X$, where $Y$ is a set of
indices identifying the subsets.  In the example of the prepartitions for
the NPP, $Y$ is the set all prepartitions, and for a prepartition $y\in Y$,
$\varphi(y)\subseteq X$ specifies the subset of solutions defining
prepartition $y$, and therefore also a particular coarse-graining of the
original problem.  The basic idea is now to replace the given search space
$X$ by the space $Y$ of restrictions, and to consider the collection of
restricted problems $\hat f_y: \varphi(y)\to\mathbb{R}$ with $\hat
f_y(x)=f(x)$ for all $x\in\varphi(y)$. In the example of prepartitions, we
have $x\in\varphi(y)$ if and only if $x_i=x_j$ when $y_i=y_j$.  In order to
ensure that the collection of restricted problems still contains the ground
state of the original problem, we require that $\bigcup_{y\in Y}
\varphi(y)=X$, i.e., $\{\varphi(y)|y\in Y\}$ is a cover of $X$. The map
$\varphi:Y\to 2^X$ that maps an ``encoding'' $y$ to a subset
$\varphi(y)\subseteq X$ is therefore called a \emph{cover-encoding map}
\cite{Klemm:18a}.

In order to explore the relationship between the cover-encoding map and the
original optimization problem, we define the \emph{oracle function}
\begin{equation}
  F(y) \coloneqq \min_{x\in\varphi(y)} f(x)
\end{equation}
as the solution of each of the restricted optimization problems. Then
$F:Y\to\mathbb{R}$ with $y\mapsto F(y)$ is again an optimization problem
that has the same solution as the original problem, provided there is an
encoding $y^*$ with $\varphi(y^*)=\{x^*\}$ for the solution $x^*$ of the
original problem. As shown in \cite{Klemm:18a}, one can readily define a
search operator (or neighborhood relations) on $Y$, such that the landscape
$F:Y\to\mathbb{R}$ has no local optima, and also provided that adaptive
walks exist to the solution $y^*$ from every starting point $y_0\in Y$.

Of course, computing $F$ amounts to solving all restricted optimization
problems and thus is essentially equivalent to computing a solution of the
original problem. We can, however, \emph{approximate} the oracle function
by means of a heuristic $G$ that produces a ``good'', if not necessarily an
exact, solution on $\varphi(y)$. In general, therefore we have $F(y)\le
G(y)$. Since the optimal solution $x^*$ appears as an unambiguous encoding
$y^*$ with $\varphi(y^*)=\{x^*\}$ in the encoded landscape, we may assume
that $G(y^*)=F(y^*)$, since this requires only that our heuristic correctly
evaluate the fitness for a search space containing a single
configuration. With the use of a heuristic for the restricted problems, we
therefore obtain a fitness landscape of $G:Y\to\mathbb{R}$ whose global
minimum identifies the global minimum of the original problem
$f:X\to\mathbb{R}$.  What we have gained, therefore, is that since $G$ is
an approximation to the oracle function $F$, it will also, at least
approximately, have the desirable properties of $F$. In other words, we can
expect to have a few shallow traps, and that ``nearly'' monotonic paths,
from any starting point $y_0$ towards $y^*$, will exist.

We emphasise here that our focus is on a given instance, rather than, as in
statistical mechanics, an ensemble (drawn from a probability distribution
specified, for example, by the joint distribution of coupling
constants). Another difference with typical situations in statistical
mechanics is that we consider a fixed number $n$ of entities (e.g.\ spins),
and do not insist that this number tends to infinity.

In the next subsection, we provide a concrete illustration of the above
ideas by using the example of a quadratic spin glass.

\subsection{Coarse-graining spin glasses: analytical construction} 
\label{sec:spinglass_generic}

In order to illustrate the link between cover-encoding maps and
coarse-grained, smaller-sized problems, we consider the Hamiltonian of a
quadratic spin glass
\begin{equation}
f(x) = \frac{1}{2}\sum_{i,j} a_{ij} x_ix_j + \sum_i b_i x_i
\end{equation}
with $x_i=\pm1$ and $1\le i,j\le n$ with $a_{ij}=a_{ji}$. The diagonal
terms are arbitrary and do not influence the ground state. At this point we
make no assumptions on the coupling constants $a_{ij}$ and $b_i$. Our aim
here is to construct a simple generic coarse-graining scheme. In  light
of the previous section, we are in particular interested in a
coarse-graining that causes only a small change in system size.

Most naturally, this is achieved by fixing the relative orientation of a
pair $k,l$ of spins.  To expose their contribution, we partition $f$ in
the following form
$$f(x) = \frac{1}{2}\sum_{\atop i,j\ne k,l} a_{ij} x_ix_j +
                \sum_{i\ne k,l} b_i x_i + 
                \sum_{i\ne k,l} a_{ik}x_ix_k +
                \sum_{i\ne k,l} a_{il}x_ix_l + 
                a_{kl} x_kx_l + b_k x_k + b_l x_l$$
Using  $x_k^2=1$ we can rewrite the last terms as
$\sum_{i\ne k,l} (a_{ik}+a_{il}x_kx_l) x_ix_k$ and 
$(b_k+ b_lx_kx_l) x_k$. Depending on whether we fix $x_k$ and $x_l$ to
be parallel or anti-parallel, we obtain two Hamiltonians 
$$f^{\pm}_{kl}(x) = \frac{1}{2}\sum_{\atop i,j\ne l} a^{\pm}_{ij} x_ix_j + 
\sum_{i\ne l} b^{\pm}_i x_i$$
with coupling constants that are modifed exactly for the terms involving 
spin $k$. These now depend explicitly on the relative orientation of
the spins $k$ and $l$. For notational convenience, we represent the frozen
spin-pair by the orientation of $k$. We have 
\begin{equation}
a^{\pm}_{ki} = a^{\pm}_{ik} = a_{ik}+a_{il}x_kx_l
\qquad\textrm{and}\qquad
b^{\pm}_k = b_k+ b_lx_kx_l
\end{equation} 
This leaves us with two choices to freeze the relative orientations of
spins, $x_l:= x_k$ or $x_l:= -x_k$, resulting in two different Hamiltonians
with $a^+_{ki}=a_{ik}+a_{il}$, $b^{+}_k= b_k+ b_l+a_{kl}$ and
$a^-_{ki}=a_{ik}-a_{il}$, $b^{-}_k= b_k-b_l-a_{kl}$, respectively.  By
construction we have $f(x)=f^{+}_{kl}(x)+a_{kl}$ if $x_k=x_l$ and
$f(x)=f^{-}_{kl}(x)-a_{kl}$ if $x_k=-x_l$. We note for later reference that
$b_i=0$ for all $i$ implies that the fields $b^{\pm}_k$ also vanish in the
modified problem. The constant terms $+a_{kl}$ or $-a_{kl}$ do not affect
the ground state of each of the two smaller problems $f^{+}_{kl}$ or
$f^{+}_{kl}$, one of which therefore retains the original ground state. A
knowledge of the offset $\pm a_{kl}$ is necessary, however, to decide which
of them contains the ground state, even given a knowledge of the ground
states of the spin glass problems on $n-1$ spin variables for the
Hamiltonians $f^+$ and $f^-$ respectively.

In terms of encodings, the fixing of orientations $x_l=x_k$ or $x_l=-x_k$
defines a partition of the original problem, which we may therefore see as
two encodings $y_{kl}^+$ and $y_{kl}^-$ with $\varphi(y_{kl}^+)=\{x\in X|
x_l=x_k\}$ and $\varphi(y_{kl}^-)=\{x\in X| x_l=-x_k\}$.

The combination of coarse-graining steps of this type is not entirely
trivial. Suppose we first freeze $k,l$ and then $h,k$. This implies that
the relative orientations of $x_h$ and $x_l$ are also fixed. A convenient
way to keep track of pairs of spins with fixed orientation is to encode
this information in a spanning forest $W$. The connected components of $W$
indicate the sets of spins that are frozen relative to each other. Each
edge is labeled with $+$ or $-$ to indicate whether the adjacent spins are
parallel or antiparallel. For two spins in the same connected component,
the relative orientation is the product of the signs along the path
connecting them. Given a spanning forest $W$ on the set of spin variables
with $\pm$-labeled edges, the relative signs of any two spins $x_i$ and
$x_j$ in a connected component of $W$ is given by the number of $-$ labels
along the unique path from $i$ to $j$ in $W$: if there is an even number of
$-$ labels, then $x_j=x_x$, if the number is odd then $x_j=-x_i$. As a
consequence, we retain one independent spin $z_c$ variable for each
connected component $c$ of $W$. Each encoding $y$, in this case, is defined
by a $\pm$-labeled spanning forest on the spin variables. All interactions
$a_{ij}x_ix_j$ within a connected component $c$ become constant since
$x_ix_j$ equals either $x_c^2=1$ or $-x_c^2=-1$.  For every $\pm$-labeled
spanning forest $y$, therefore, we obtain again a quadratic spin glass
Hamiltonian $f_y$ that differs from $f$ by this additive factor. Every spin
state $z$ in the coarse-grained model together with the encoding $y$
defines a spin state $x_y(z)$ in the original spin glass: To obtain $x_i$
from $z_c$ for a spin in connected component $c$ of $y$, follow the path
from the representative and account for the $-$ labels. Thus we have
$f(x(z)) = f_y(z)+a_y$, where $a_y$ is the extra additive constant
accounting for the frozen interactions within each of the connected
components. The Hamiltonian $f_y(z)$ is the coarse-grained spin glass,
which is defined in terms of representative spin states $z_c$ of the
connected components $c$ of $W$. Its ground state is $z^*=\argmin_z
f_y(z)=\argmin_z f(x(z))$, where the second equality holds because $a_y$ is
a constant that, however, depends explicitly on the encoding $y$. The spin
vector $z^*$ can be translated back to the original system as follows: for
each $z$, the $\pm$-labeled spanning tree $W$, i.e., the encoding $y$,
uniquely determines the spin vector $x(z)\in\varphi(y)$. Therefore we can
write the ground state of the original problem restricted to $\varphi(y)$
as $x^*\coloneqq x(z^*)=\argmin_{x\in\varphi(y)} f(x)$.  Together, the
coarse-grained spin glass models therefore determine the oracle function
\begin{equation} 
  F(y)=f(x^*)=f_y(z^*)+a_y\,.
  \label{eq:oracle+ay}
\end{equation}
While the constants $a_y$ do not matter as far as the optimization on the
coarse-grained Hamiltonians $f_y$ is concerned, they cannot be neglected
at the level of the oracle function, as well as its approximation in
practical computations. 

The spanning-forest-based construction outlined above can be specialized
and further simplified in several ways. For example, the hierarchical block
spin RG procedure \cite{Kadanoff:66} can be seen as a special case in which
the only spanning forests considered are those whose connected components
are blocks of spins with mutually frozen orientations.  The hierarchical
aggregation of blocks then corresponds to a hierarchy of spanning forests,
with sub-forests which in turn correspond to finer aggregations.

\subsection{Coarse-graining spin glasses:
  computational example of a spin glass encoding}

\begin{figure}
\includegraphics[width=\textwidth]{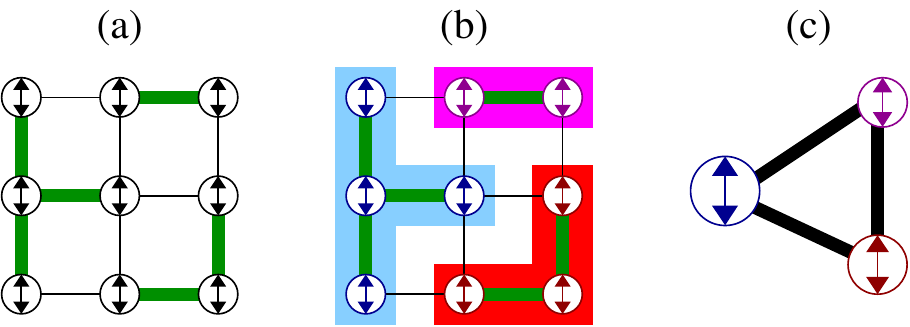}
\caption{
Illustration of the encoding by spanning forests on a spin glass system
with $3 \times 3$ sites (circles with up-down arrows). (a) A spanning forest
drawn with thick lines as a subset of the bond set. (b) Each connected
component of the spanning forest induces a cluster (block) of spins, with
blocks shown as shaded areas. (c) Mutual orientations of spins inside a
block are fixed as parallel or antiparallel. Here, for simplicity and as
done in the simulations, all edges of the forest are taken as labeled $+$.
Spins inside a block are thus mutually parallel, combining into a single
block spin each. At difference with the actual systems simulated, the
square grid in this illustration is non-periodic. 
}
\label{fig:spin_forest}
\end{figure}

As a case study, we consider a quadratic spin glass albeit with vanishing
terms $b_i=0$. We consider spins arranged on an $L\times L$ square grid
with periodic boundary conditions. We label the vertices $N_\rfloor =
\{1,2,\dots,N\}$ and write $B$ for the bond set of this square grid.  For
sites $i,j \in N_\rfloor$, the initial coupling strength $a_{ij} \neq 0$
only if $\{i,j\} \in B$. Then $a_{ij} \in [-1;+1]$ is drawn uniformly and
independently.

We use a variation of the encoding described in the previous section and
lump together parallel spins only. Since the spanning forest itself serves
as the encoding, we simply write $y$ also for the spanning forest.  Every
connected component $c$ of the spanning forest $y$, by assumption,
satisfies $x_i=x_j=z_c$ for all $i,j\in c$.  The set $Y$ of encodings thus
coincides with the set of (unlabeled) spanning forests. We note that two
spanning forests $y$ and $y'$ with the same connected components satisfy
$\varphi(y)=\varphi(y')$ and thus also $F(y)=F'(y)$. However, partitions
and spanning forests suggest different move sets, i.e., neighborhood
relations on $Y$. We also note that, for a given $y$, we have
\begin{equation}
  \begin{split}
  f(x(z)) &= \sum_{c,d\in C_y, c \neq d}  
  \underbrace{\left(\sum_{i\in c, j\in d} a_{ij}\right)}_{\eqqcolon a^{(y)}_{cd}} z_cz_d + 
  \underbrace{\left(\sum_{c \in C_y} \sum_{i,j\in c} a_{ij} \right)}_{\eqqcolon \bar a_y}\\
  &=\sum_{c,d\in C_y} a^{(y)}_{cd} z_cz_d + \bar a_y
  = f_y(z) + \bar a_y
  \end{split}
\end{equation}
Here, $C_y$ denotes the set of connected components of the spanning forest
$y$ and $z_c$, $z_d$ are the spin orientations for components $c$ and $d$,
respectively. Since all spins have the same orientation $z_c$ within a
component $c$, each component also contributes a constant. Each $y$,
therefore, gives rise to a smaller quadratic spin glass problem $f_y(z)$
and an offset $\bar a_y$ that accounts for the interactions within
connected components.

As noted, the oracle function $F(y)$ cannot be computed efficiently in
practice.  We use the following procedure as a heuristic approximation to
optimize the restricted problem $f_y(z) = \sum_{c,d} a^{(y)}_{c,d} z_c
z_d$, i.e., to compute the approximation $G(y)$ of the oracle function
$F(y)$.

\begin{algorithm}
  \caption{Approximate Oracle Function $G(y)$}
  \label{alg:oracle}

  \KwData{$a^{(y)}_{c,d}$ for all connected components $c,d$ of $y$} 
  Initialize $a'_{c,d} \leftarrow a^{(y)}_{c,d}$ for all connected 
  components $c,d$ of $y$\;
  \While{more than two spin variables (connected components) are left}{
    $\{p,q\}\leftarrow \argmin a'_{p,q}$ \;
    unify components $p$ and $q$\;
    $a'_{r,p\cup q}\leftarrow a'_{r,p}+a'_{r,q}$\;
  }
  $\hat x\leftarrow x(\hat z)$ is the spin vector corresponding to the
  bipartition of $N_\rfloor$ defined by $\hat z$ on the remaining two
  connected components\;
  $G(y)\leftarrow f(\hat x)$\;
\end{algorithm}

The idea behind this heuristic is that a bond with the largest (positive)
coupling strength is likely to connect two spins with equal orientation in
the ground state. The final ingredient is a move set on the forests.  To
this end we stipulate that two spanning forests are adjacent if they differ
by a single edge. Note that insertion (or deletion) of an edge decreases
(or increases) the number of connected components by one. Since both $y$
and $y'$ are spanning forests by assumption, only edges between connected
components can be inserted.
  
Initialized with the empty forest $y(0)=\{\}$, an adaptive walk in the
encoded landscape $(Y,G)$ thus iterates the following operations. Draw $i,j
\in N_\rfloor$ uniformly. If $i$ and $j$ are in different connected
components of the forest $y(t)$, set $\tilde y=y(t)\cup\{\{i,j\}\}$;
otherwise select an edge $\{k,l\} \in y(t)$ uniformly at random and set
$\tilde y=y(t) \setminus \{\{i,j\}\}$; if $G(\tilde y)\le G(y(t))$ set
$y(t+1)=\tilde y$, otherwise set $y(t+1)=y(t)$. Thus in each time-step we
generate a proposal $\tilde y$ by random addition or removal of an edge; we
accept $\tilde y$ if, evaluated by $G$, it is not worse than the current
solution $y(t)$.

For comparison, we consider adaptive walks on the direct landscape. Here we
choose an initial spin vector by drawing $x(0)$ from $\{-1,1\}^N$
uniformly. Then in each time step we perform the following operations in
each time step. Generate a proposal $\tilde x$ by flipping spin $i$ in
$x(t)$ with $i \in N_\rfloor$ selected uniformly at random. If $f(\tilde x)
\le f(x(t))$ set $x(t+1) = \tilde x$, otherwise set $x(t+1) = x(t)$.

\begin{figure}
\includegraphics[width=\textwidth]{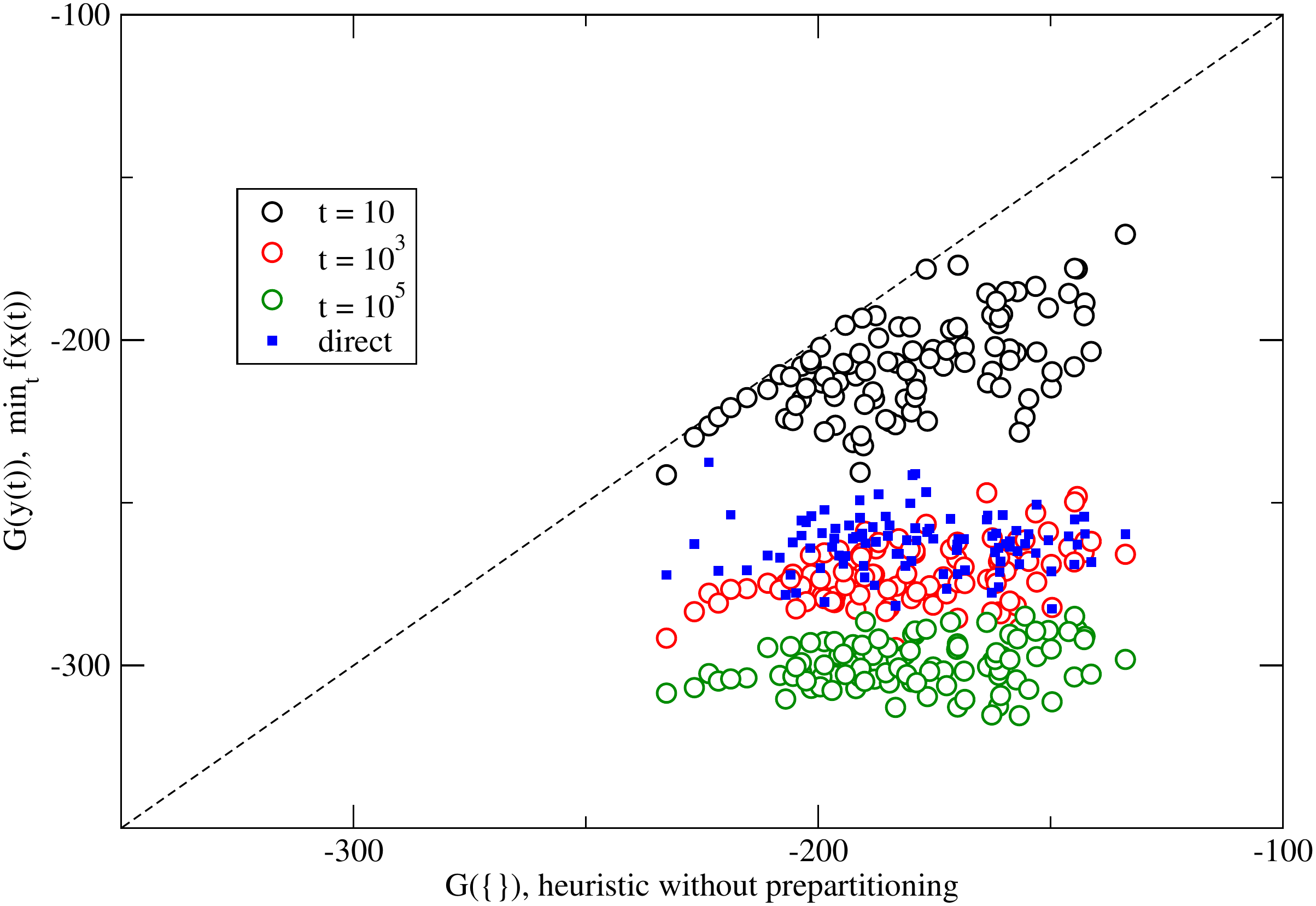}
\caption{Results of adaptive walks on the landscapes of spin glasses on a
  $20 \times 20$ square grid. Each plotted point is the result of an
  adaptive walk on a randomly generated spin glass (see main text). Open
  circles show the value of the approximation $G(y(t))$ at times $t \in
  \{10, 10^3, 10^5\}$ versus G(\{\}), the value of the approximation
  without prepartitioning. Time $t$ is measured in elementary updates of
  $y$, i.e.\ addition or removal of a single edge from the spanning
  forest. Each walk on the encoded landscape is initialized with the
  edge-less forest $y(0)=\{\}$, i.e., partitioning into singletons. For
  comparison, the minimum of the cost function $f(x(t))$ for $t \rightarrow
  \infty$ is shown for adaptive walks in direct encoding (filled squares)
  on the same spin glass instances.}
\label{fig:glassa_0}
\end{figure}

Figure \ref{fig:glassa_0} shows that the adaptive walks in the encoding
outperform those on the direct landscape after a sufficiently long time, $t
\approx 10^5$. The heuristic alone without prepartitioning, however,
performs worse than the adaptive walks on the direct landscape. This is
different from the other encoding scenarios we have encountered so far
\cite{Klemm:18a}. For the number partitioning problem, for instance, the
Karmarkar-Karp heuristic \cite{Karmarkar:83} alone outperforms the adaptive
walks on the direct landscape. Here, the combination of the heuristic with
the prepartioning encoding is required for a performance gain.

In order to assess the significance of the performance gain observed
in Figure \ref{fig:glassa_0}, we compare the energy $\eta_{\min}$ reached
by an adaptive walk of length $t$ in the encoded landscape
$G: Y\to\mathbb{R}$ with the distribution of energy values $E$
reached by a sample of adaptive walks with restart 
of the same length $t$ in the direct landscape $f:X\to \mathbb{R}$.

Local search with restart serves as a heuristic solver for combinatorial
optimization problems \cite{Selman:92,Mengshoel:11}. After a given number
of steps $\tau$, the current solution is replaced by a spin configuration
$(x_1,\dots, x_N)$ drawn uniformly at random and the walk is continued
from there. With total run time $t$ being a multiple of $\tau$, this
amounts to independently running $t / \tau$ adaptive walks from different
initial conditions and observing the overall minimum energy
$\eta_{\min}$ encountered.

\begin{figure}
\includegraphics[width=\textwidth]{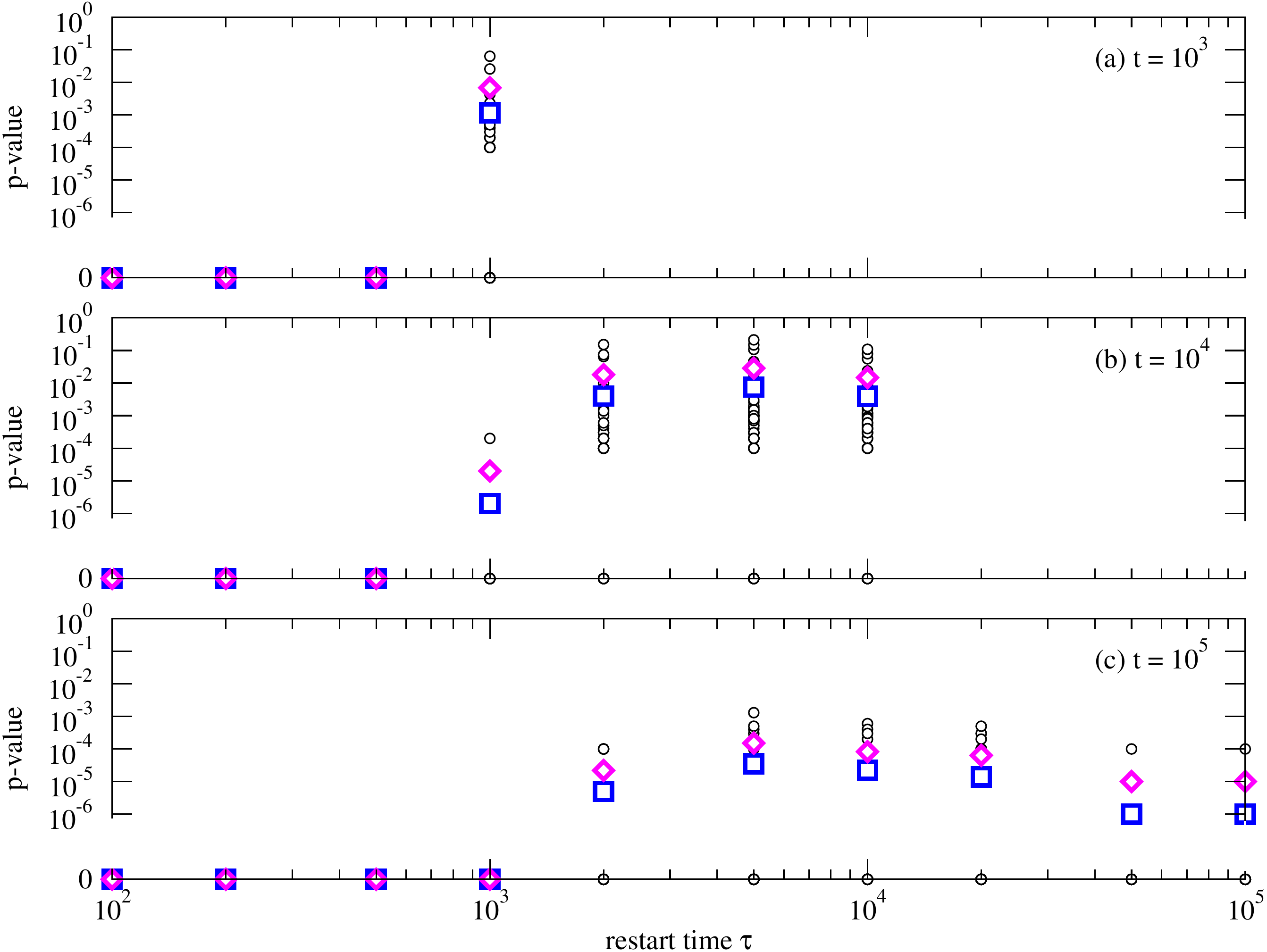}
\caption{Performance statistics on 100 randomly generated spin glasses with
  $20 \times 20$ sites (same as in Figure~\ref{fig:glassa_0}). On each
  encoded landscape, a single adaptive walk of $t$ steps is performed with
  the resulting minimum energy denoted by $\eta_{\min}$. The p-value
  (plotted as small circle) is the fraction of adaptive walks with restart
  in the direct (single spin-flip) landscape encountering an energy value
  $E \le \eta_{\min}$, also within $t$ steps. The mean (squares) and
  standard deviation (diamonds) over the 100 spin-glass realizations are
  shown for walk lengths $t \in \{10^3,10^4,10^5\}$ and
  restart time $\tau \le t$.  Each p-value is based on 10000 adaptive walks
  with restart in the direct landscape. The ordinate axes is split into
  a part with logarithmic scale and a part displaying the value zero.}
\label{fig:glassa_pval}
\end{figure}

In this setting, the empirical p-value for an improved performance of
the encoding is simply the fraction of direct adaptive walks reaching
$E\le\eta_{\min}$. Figure~\ref{fig:glassa_pval} summarizes the results
for the single spin-flip landscape for multiple spin-glass realizations
(to estimate the variance) and different walk lengths $t \in
\{10^3,10^4,10^5\}$ (panels (a), (b), (c)) and restart time $\tau \le t$.
The majority of p-values are very small, and decrease as the walk length
increases. For walk length $10^5$, panel (c), all p-values are below
$2\times 10^{-3}$. We conclude that the encoded walks outperform all
direct walks in each spin glass.

\begin{figure}
\includegraphics[width=\textwidth]{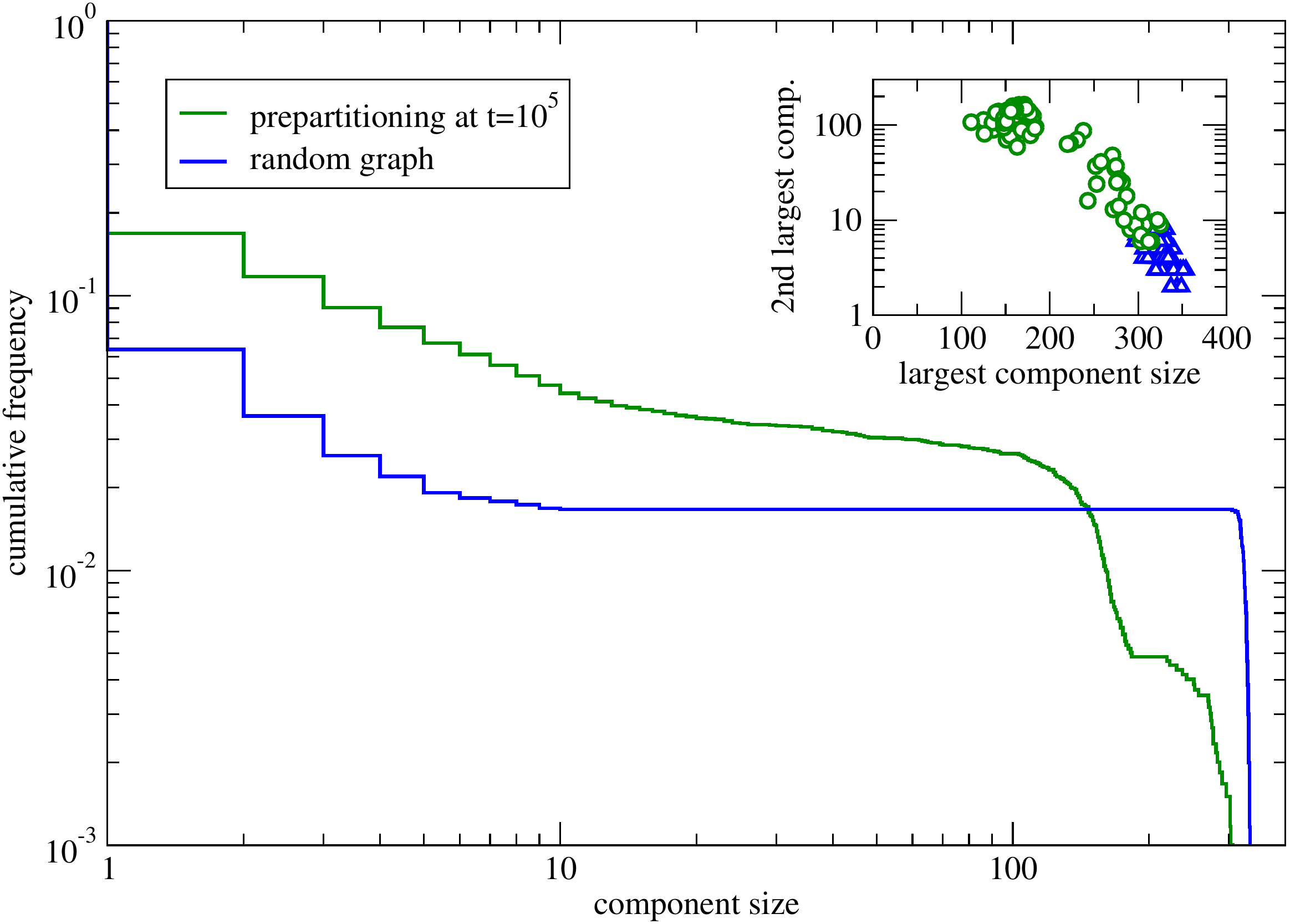}
\caption{Main panel, green curve: cumulative
  distribution of cluster (connected component) sizes of adaptive walks in
  the encoded landscape at time $t=10^5$, taken over the same 100 instances
  as in Figure~\ref{fig:glassa_0}. For surrogate spanning forests with the
  same number of nodes and edges, the blue curve is the cluster size
  distribution. In the inset, the largest vs. the second largest cluster
  size is shown for each of the 100 instances, again for the actual
  prepartioning (green circles) and for random surrogate forests (blue
  triangles).}
\label{fig:glassacl_0}
\end{figure}

We analyze the prepartioning in terms of the distribution of connected
component sizes as shown in Figure \ref{fig:glassacl_0}. Compared to random
forests with the same number of edges and nodes, the prepartitions
generated by the encoded adaptive walks have a lighter tail but exhibit an
additional peak at larger cluster sizes. The inset of
Figure~\ref{fig:glassacl_0} shows a salient difference between actual
prepartitions and surrogate forests: In most instances of adaptive walks,
the second largest cluster of the prepartitioning is roughly of the same
size as the largest cluster, an effect not seen in the surrogate forests
where the second largest cluster is of negligible size.

We end this section by considering the computational complexity of
cost function evaluation. In the direct landscape, the cost
function $f:X\to \mathbb{R}$ is a sum over $|B|$ terms, one for each bond
of the spin glass. The time complexity of evaluating $f$ is
thus ${\mathcal O}(|B|)$. In an adaptive walk, however, only the
difference of cost values between two spin configurations disagreeing
at a single site $i$ needs to be calculated. Then merely the terms
for bonds incident on site $i$ are evaluated which leads to constant
time complexity ${\mathcal O}(1)$ if the number of neighbours is upper
bounded independently of $|B|$. For the encoded landscape, evaluation
of the cost function $G: Y\to\mathbb{R}$ has time complexity
${\mathcal O}(|B|)$ as well. This is due to the fact that each of the
$|B|$ bonds is eliminated or lumped together with another bond at most
once, either in the initial cluster formation by the prepartitioning,
or in the heuristic Algorithm~\ref{alg:oracle}. Also here in the context
of an adaptive walk, computing the cost difference between two adjacent
encodings may require time scaling merely sublinearly with $|B|$ where 
details are subject to further analysis. The space complexity
is ${\mathcal O}(|1|)$ for evaluating $f:X\to \mathbb{R}$ and
${\mathcal O}(|B|)$ for evaluating $G: Y\to\mathbb{R}$.

\section{Formal analogies between cover encodings and the
  renormalisation group }

\subsection{Semigroup structure of cover-encodings} 

It is tempting to speculate that the coarse-grainings we have observed in
the above are analogous to those observed in renormalization group theory,
well known for its use in analyzing spin glasses and related disordered
systems \cite{Rosten:12}. This intuition can be expressed more precisely as
follows: For a given type of problem, such as the NPP or the TSP, consider
the space $\mathfrak{X}$ of all possible instances of all sizes. A
particular instance is a point $\mathbf{x}\in\mathfrak{X}$. Now we define a
set $\mathcal{R}$ of maps $r:\mathfrak{X}\to \mathfrak{X}$ that map larger
instances to strictly smaller ones. Of interest in this context are in
particular those maps $r$ that (approximately) preserve salient
properties. Note that since $r(\mathbf{x})$ is a smaller instance than
$\mathbf{x}$, the map $r$ is not invertible -- several larger instances can
be coarse-grained to the same smaller one. Also, $\mathcal{R}$ can be
composed, (i.e., if $\mathbf{x'}=r_1(\mathbf{x})$ and
$\mathbf{x''}=r_2(\mathbf{x'})$, then $\mathbf{x''}=r_2(r_1(\mathbf{x}))$),
so that $\mathbf{x''}$ is a valid coarse-graining of $\mathbf{x}$. The
coarse-grainings thus form a semi-group which is reminiscent of the
\emph{renormalization group} (RG) \cite{Wilson:74,Wilson:71}.

As mentioned above, a technical problem arises from the innocuous-looking
notion of ``the space $\mathfrak{X}$ of all possible instances of all
sizes''. While this is harmless e.g.\ for the NPP, where a coarse-graining
of an instance with $n$ numbers, $\mathbf{x}=\{a_1,a_2,\dots,a_n\}$, is
again an NPP on a smaller set of numbers that are partial sums of the
original ones, problems can arise for other systems.  Contracting a pair of
spins in a lattice-based spin glass, for example, destroys the overall
configuration; while $r(\mathbf{x})$ is a meaningful coarse-graining of
$\mathbf{x}$, it no longer belongs to the problem class that we started
with. This can be remedied in two ways: (i) we may restrict ourselves to
coarse-grainings that guarantee $r(\mathbf{x}) \in \mathfrak{X}$ for all
$\mathbf{x}\in\mathfrak{X}$ and all $r\in\mathfrak{R}$, or (ii) we give up
the notion of coarse-graining in the sense of creating a smaller problem
instance, and instead consider coarse-graining as the imposition of
restrictions on the solution space.

\subsection{Cover-encodings viewed as generalized coarse-graining schemes} 

The coarse-graining scheme in Sect.~\ref{sec:spinglass_generic} turned out
to be closely related to the cover-encoding maps. The key observation for
our purposes is that the restriction of $f$ to $\varphi(y)$ serves as a
slightly generalized version of the coarse-grained problem $f_y$. A series
of additional examples of cover-encoding maps can be found in
\cite{Klemm:18a}. Some of these correspond to obvious coarse-graining
schemes, while for others the cover-encoding leads to smaller instances of
modified problems. The close relationship of cover-encoding maps to
coarse-graining schemes, together with the technical difficulties arising
from direct coarse-grainings, are reminiscent of the renormalisation group;
we are thus prompted to explore analogies between cover-encoding maps and
RG.

A simple example of cover-encoding maps are the so-called \emph{schemata}
\cite{Altenberg:95a}, which are frequently used e.g., in the analysis of
genetic algorithms.  Assuming $X\coloneqq \{0,1\}^N$, the set of binary
strings of length $N$, the corresponding schema encoding uses string $y\in
\{0,1,*\}^N\eqqcolon Y$ as encoding variables. A string $x$ belongs to the
schema if $y$ coincides with $x$ on all fixed positions, and has either a
$0$ or a $1$ at the positions where $y$ shows a wildcard character $*$.
Thus
\begin{equation}
  \varphi(y) = \left\{x\in X| x_i=y_i \text{ if } y_i\in\{0,1\} \right\}
\end{equation}
Note that $\varphi(y)=\{y\}$ if all positions in $y$ are fixed, i.e., if
$y$ contains no wildcard, and $\varphi(y)=X$ if $y$ consists of wildcards
only. Schemata have been used for the RG analysis of models of genetic
dynamics, see e.g.\ \cite{Stephens:02,Stephens:04,Stephens:09}. This
connection provides yet another motivation to search for a possible
connection between cover-encoding maps and RG formalisms.

The second key ingredient is the symmetric adjacency relation $\sim$ on
$Y$, which establishes a proximity relation on $Y$, the space of encoding
variables.  In \cite{Klemm:18a} we defined $\sim$ independently of
$\varphi$. Neverthess, the idea is that $y'\sim y''$ should be closely
linked with properties of $\varphi$, i.e., $y'\sim y''$ should be true if
$\varphi(y')$ and $\varphi(y'')$ are similar.

Moving from $y'$ to an adjacent $y''$ may correspond to both
coarse-graining and fine-graining. To introduce a directionality, we
therefore need a measure for the number of degrees of freedom in
$\varphi(y)$. For simplicity, we use here the cardinality $|\varphi(y)|$,
but any measure $|\,.\,|$ that is monotonic in the sense that
$\varphi(y)\subsetneq\varphi(y')$ implies $|\varphi(y)|<|\varphi(y')$ could
be used. In the case of schemata, for example, the number of degrees of
freedom is simply the number of wildcards, i.e., $\log_2 |\varphi(y)|$. In
our spin glass example, the number of spin variables appearing in the
Hamiltonian $f_y(z)$ is also $\log_2 |\varphi(y)|$. Similar monotonic
relations hold for all examples described in \cite{Klemm:18a}.

Step-wise coarse-graining is thus achieved by moving from $y'$ to an
adjacent $y''$ such that $|\varphi(y')|\ge |\varphi(y'')$. This turns
$(Y,\sim)$ into a directed graph $\Gamma(Y,E)$, replacing the undirected
edges by a set $E$ of arcs pointing towards neighbors with a non-increasing
number of degrees of freedom. For technical reasons, we include equality
here so as to connect more directly to the formalism of 
\cite{Klemm:18a}. There, ``sideways'' steps were required in some examples
to ensure the ergodicity of the neighborhood relation $\sim$.

\subsection{The Flow Semigroup}

In the digraph $\Gamma(Y,E)$ we consider, for every arc $e=(y,y')\in E$ the
so-called \emph{elementary collapsing map} \cite{Howie:66,Howie:78}
defined by
\begin{equation}
      T_e(v) = T_{(y,y')}(v) = \begin{cases}
             y' & \text{if } v= y \\
             v & \text{otherwise}
             \end{cases}
\end{equation}
That is, $T_{(y,y')}$ maps $y$ to $y'$ and leaves every other vertex
invariant. Hence, $T_e$ is idempotent, i.e., $T_e(T_e(v))=T_e(v)$ for all
$e\in E$. The elementary collapsing maps generate the transformation
semigroup $\mathcal{S}(\Gamma)=\langle T_e | e\in E\rangle$ acting on $Y$,
which is known as the \emph{flow semigroup} or \emph{Rhodes semigroup} of
$\Gamma$ \cite{Rhodes:10}. It can be shown that
$T_{(y,y')}\in\mathcal{S}(\Gamma)$ if and only if $(y,y')\in E$ or $(y',y)$
is contained in a directed cycle of $\Gamma$ \cite{Yang:06}. The flow
semigroup therefore completely determines the undirected graph underlying
$\Gamma$ \cite{Yang:09,Rhodes:10,Horvath:17}. The relationships between the
properties of $\Gamma$ and properties of $\mathcal{S}(\Gamma)$ are explored
in some more detail in \cite{East:17}.

Now consider a function $h$ on $Y$ that is non-increasing along all edges
of $G(Y,E)$. In particular, our measure for the degrees of freedom, i.e, $h:
y\mapsto|\varphi(y)|$, has this property. Then $h(T_e(v))=h(y')\le h(v)$ if
$v=y$ since $h$ is non-increasing along the directed edge $(y,y')$, and
$h(T_e(v))=h(v)$ otherwise. Any concatenation $T$ of elementary collapsing
maps, i.e., any $T\in\mathcal{S}(G)$ thus satisfies $h(T_e(v))\le h(v)$ for
all $v\in Y$.  The flow semigroup $\mathcal{S}(G)$ on $G(Y,E)$ therefore
encapsulates a key property of the coarse-grainings: The action of every
$T\in \mathcal{S}(G)$ on $y$ produces an encoding $Ty$ such that
$|\varphi(Ty)|\le |\varphi(Ty)|$, i.e., the number of degrees of freedom is
non-increasing. 

The semi-group formalism also provides a convenient way of expressing the
reachability conditions imposed in \cite{Klemm:18a}. Most naturally one
would require:
\begin{itemize}
\item[(R)] If $\varphi(y')\subseteq \varphi(y)$ then there is a $T\in
  \mathcal{S}(G)$ such that $y'=T(y)$. 
\end{itemize}
Condition (R) simply says that whenever $y'$ is a proper coarse-graining of
$y$ in the sense that $y'$ encodes a restriction of $\varphi(y)$, then
there is a directed path in $\Gamma(Y,E)$ from $y$ to $y'$ such that the
number of degrees of freedom is non-increasing along the path. This
interpretation follows directly from the definition of $\mathcal{S}(G)$
since $T\in\mathcal{S}(\Gamma)$ if and only if there is a sequence of edges
$e_1,\dots,e_k$ such that $T=T_{e_k}(T_{e_{k-1}}(\dots(T_2(T_1))\dots))$.
In other words, condition (R) ensures that stepwise coarse-graining along
the edges of $\Gamma(Y,E)$ can be concatenated in such a way that every
proper coarse-graining can be achieved.

\subsection{Overview of relationships with the renormalization group} 

Cover-encoding maps and the associated flow semigroup provide a useful
\emph{mathematical substrate} to construct, in a principled manner,
coarse-graining schemes that may be amenable to RG analysis. However, the
formalism does not provide in itself the quantities that are to be
considered, i.e., those from which criticality conditions are derived.

It is interesting to note that cover-encoding maps can also be combined
with coordinate transformations, i.e., bijective maps $\xi:X\to
X'$. Instead of the original representation, one can first transform the
given Hamiltonian/cost function to a different basis, and then construct a
cover-encoding map for the transformed coordinate system. For the discrete
models at hand, Walsh transformations \cite{Goldberg:89a} or the use of
characters instead of group elements \cite{Rockmore:02a} are natural
choices corresponding to discrete versions of Fourier
transforms. Occasionally also other coordinate systems, e.g., Haar
transforms, have been advocated in this context \cite{Khuri:94}. The
possibility of making use of these types of coordinate transformations
suggests that our cover-encoding framework may well be relatable to
$k$-space RG.

As mentioned in a previous section, there is a major difference between the
use of RG in statistical physics and the cover-encoding framework outlined
above. While RG is primarily concerned with the rescaling of the
distributions of observables in large systems to determine their critical
behaviour, cover-encodings are largely concerned with single instances of
arbitrary size. In practical applications, our flows on
$\mathcal{S}(\Gamma)$ are determined by a stepwise decrease of $G(y)$, the
approximation of the elusive oracle function, in the form of an adaptive
walk. In the RG framework, by contrast, one considers continuous flows that
can often be studied analytically.

This does not yet rule out an analogy, assuming that something like an
ergodic hypothesis applies, where the behaviour of typical instances is
indeed that of the average. Thus, starting from $\mathbf{x}=(X,f)$, or more
precisely, an encoding $y$ so that $\varphi(y)=\mathbf{x}$, we can think of
adjacent encodings $y'\sim y$ with $|\varphi(y')|<|\varphi(y)|$ as
``renormalized'' versions of $\varphi(y)$. A path in $(Y,\sim)$ leading
from $\mathbf{x}$ to the trivial instance thus can be seen as the iteration
of progressively renormalized samples.

A positive example of this analogy could be that of the spanning forest
encoding of the NPP with real-space renormalization schemes for Ising
spins: an example of an $\mathcal{R}$ could be a so-called block spin
transformation \cite{Kadanoff:66}, where suitable averages are taken over
small local subsets of spins, which are then progressively scaled up to
larger system sizes to explore their critical behaviour.  Only certain
block variables will work for such schemes, depending on the underlying
symmetries of the problem, just as, in the earlier subsection, only the
sums of numbers $a_i$ preserve the optimal solutions.  Such simple
real-space scalings, do not, however, always exist for our optimization
schemes: the prepartition encoding of the TSP, for example, cannot be
rephrased as a coarse-grained (i.e., reduced-size) TSP. To see this, simply
observe that the evaluation of a tour in the restricted model still
requires an optimization over multiple incoming and outgoing connections
(roads) for every city, i.e., the information of inter-city distances
cannot be collapsed in any way upon the transition from a larger (less
restricted) to a smaller (more restricted) problem. This does not, however
rule out the possibility of, say, a renormalisation-type scaling in some
sort of generalised Fourier space. In the case of landscapes on permutation
spaces, the characters of the symmetric group provide a suitable
Fourier-like basis \cite{Rockmore:02a}, which seem to be applicable to TSP
and certain assignment problems.

\section{Discussion} 

In this paper we have examined the ideas behind our cover-encoding schemes
presented in \cite{Klemm:12b,Klemm:18a} to highlight the role of
coarse-graining in them.  The basic need for cover-encoding schemes arises
because search algorithms typically encounter local minima in their quest
for the perfect, globally optimal solution. The cover-encodings method of
resolving this problem is to partition the search space (either in its
original, or in a transformed, form) into segments (encodings), such that
the set of these partitions also contains the globally optimal solution
(i.e., so that the encodings are now cover-encodings). While an assumed
'oracle' is able to guide us to the perfect ground state, its elusiveness
leads us into a process of educated guesswork, whereby we choose the best
available ``approximate oracle'' that can guide us to the lowest energy
states amenable to search in a reasonable time. In our earlier work
\cite{Klemm:12b,Klemm:18a}, we have given concrete examples of how to
find and apply such approximate oracles to NP-complete problems.

We have also presented a study of the spin glass landscape encoded with
prepartitioning by spanning forests. The heuristic employed is relatively
inefficient in the sense of being outperformed by local search on the
direct (single spin flip) landscape when applied alone, i.e. without
prepartitioning.  The scenario implies a performance ranking which is
different from previous examples. The study \cite{Ruml:1996} of the number
partitioning problem with the largest differencing heuristic
\cite{Karmarkar:83} and encoding by prepartioning reveals a performance
ranking:
\begin{equation*}
  \text{direct landscape search} < \text{heuristic} < \text{encoded
    landscape search}
\end{equation*}
For the spin glass encoding studied here, we observe that the
prepartitioning evolves towards arranging the set of spins in two large
clusters, that together nearly span the whole system, therefore being close
to providing an actual solution.  The heuristic is left with local
processing, merging each of the remaining smaller clusters into one of the
two large ones. A statistical comparison of the performance
  differences nearly unambigously favours encoded adaptive walks over
  the direct landscape, suggesting suitable cover encodings as an appealing
  approach to algorithm design.

The main purpose of this contribution, however, is a better
  understanding of the similarity of cover-encoding schemes with those of
the renormalisation group.
\begin{itemize}
\item Cover-encoding schemes can be used to provide (approximate) solutions
  of particular instances of problems. To this end, they require the
  approximate solution of a large number of (typically) smaller
  optimization problems. This is most useful if the original problem is
  NP-hard. Since computing the oracle function is (at least) as difficult
  as solving the original problem, the practical advantage derives from
  replacing $F(y)$ by an approximation $G(y)$ that can be computed more
  efficiently. Clearly, the use of cover-encodings makes no sense if
  $\min_{x\in X} f(x)$ can be computed efficiently.  Similarly, while
  renormalization group schemes can be used for a wide range of systems,
  they are typically of the greatest interest for complex problems where
  exact solutions are unavailable; in these cases, RG is applied to provide
  approximate solutions of small and typical systems, with the
  understanding that these will be scalable to the thermodynamic limit.
  We have suggested that in the presence of ergodicity, the solution of
  ''particular'' instances in the cover-encoding optimisation schemes may
  be translatable to that of ''typical'' instances in statistical
  physics. We have also established that our cover-encoding schemes have
  the semigroup property, as is needed for a stricter analogy with
  renormalisation group schemes.

\item The coarse-graining underlying our cover-encoding schemes can be
  applied, as shown in the body of the paper, to situations where degrees
  of freedom are progressively frozen out and replaced by far fewer
  ''block'' degrees of freedom, such as in the spanning forest encoding of
  the TSP. This is strongly analogous to the method of block spins in the
  renormalisation group \cite{Kadanoff:66}. Sometimes, however, the
  coarse-graining picture for our cover-encoding schemes is more subtle --
  it may not be in real space, but in some more complex space, e.g.\ as was
  seen in the prepartition approach to the TSP mentioned above. This bears
  at least a superficial resemblance to $k$-space renormalisation schemes,
  where it is difficult to visualise in real space the scale invariance
  between original and renormalised configurations.

\item Essential features of the cover-encodings oracle \cite{Klemm:18a}
  include a high degree of redundancy, an adjacency structure and a path to
  the global optimum. An interesting analogy in $k$-space RG concerns its
  smoothing schemes: faced by a divergence and/or a mismatch between
  original and effective variables, the path of an RG trajectory is
  re-defined on a smoother path to its fixed points using regularity
  schemes, or the ignoring of higher-order terms in Taylor expansions.

\item Coping with strong disorder, in both cover-encoding as well as RG
  schemes, involves ''averaging out'' disorder; this is done in RG using,
  say, replica-symmetric (RSB) methods \cite{mpv,riccardo}, as well as
  being implicit in the choice of 'approximate' oracles for, say, the
  quadratic spin glass problem examined in the present paper.

\item In both cover-encoding and RG methodologies, there is an element of
  choice involved: the wrong choice of trajectory in an RG scheme can get
  one to a trivial fixed point and equally, the wrong choice in an
  optimisation scheme can get one to a solution that is far from the
  optimal one. The choice of the correct trajectory is, in the case of
  optimisation schemes that are not fully analytic, based on trial and
  error and the use of computer algorithms; this is also the case for RG,
  where in \cite{jerome}, for example, the use of a greedy algorithm was
  combined with renormalisation group ideas to find an optimal solution to
  the TSP.
\end{itemize}

What we have presented above is a list of rather compelling analogies
between optimisation schemes using cover-encodings, and those based on the
ideas of the renormalisation group. These are not by any means
complete. Questions regarding the details of the flow dynamics in both
schemes remain. Let us assume that trivial/critical solutions in RG
correspond to suboptimal/optimal solutions for encodings. In RG schemes,
one assumes ''linearisability''/smoothness around fixed points and the
classification of adjoining trajectories as being stable or unstable. We
await a more rigorous extension of such ideas to cover-encoding methods,
which might make it easier to identify those paths that lead close to the
global minimum.

We emphasise in conclusion that the analogies referred to above do not, in
fact, address the most difficult enigma in either scheme: that of the
choice of the appropriate encoding (cover-encoding schemes) or of the
appropriate scaling method (RG schemes).  In particular, we recall that of
the four encodings applied to the Number Partitioning problem
\cite{Ruml:1996}, the prepartitioning encoding turned out to be the most
efficient; while this appears to be because of the introduction of
diversity (involving the shuffling of numbers globally as well as locally),
this has not been rigorously proven. In the same way, optimal RG schemes
for complex systems involve methodologies that involve science as much as
art in their choice. We hope that our present paper provides seed material
for further investigations that may resolve these enigmas.

\section*{Acknowledgements}
  This work is supported in part by the European Research Council (ERC) under
  the European Union's Horizon 2020 research and innovation programme
  (grant agreement N.\ 694925) and the German Federal Ministry of Education
  and Research BMBF through DAAD project 57616814 (SECAI, School of
  Embedded Composite AI). KK acknowledges financial support from Project 
  PID2021-122256NB-C22 funded by MCIN/AEI/10.13039/501100011033 / FEDER, UE.

\section*{References} 
  
\bibliographystyle{unsrt} 
\bibliography{enc3}

\end{document}